\pdfoutput=1

\documentclass[11pt]{article}
\usepackage[dvipsnames]{xcolor}

\usepackage[]{acl}

\usepackage{times}
\usepackage{latexsym}

\usepackage[T1]{fontenc}

\usepackage[utf8]{inputenc}

\usepackage{microtype}

\usepackage{colortbl}
\usepackage{appendix}
\usepackage{makecell}
\usepackage{booktabs}
\usepackage{arydshln}
\usepackage{amsmath}
\usepackage{graphicx}
\usepackage{multirow}
\usepackage{cleveref}
\usepackage{enumitem}
\usepackage{amsfonts}
\usepackage{subcaption}
\usepackage{algorithm}
\usepackage{algorithmicx}
\usepackage[noend]{algpseudocode}
\usepackage{soul}
\crefname{section}{\S}{\S\S}
\Crefname{section}{\S}{\S\S}
\crefname{appendix}{Appendix}{Appendices}
\Crefname{appendix}{Appendix}{Appendices}
\DeclareMathOperator*{\argmax}{arg\,max} 
\definecolor{hzw}{RGB}{223, 97, 76}

\sethlcolor{Melon}

\newcommand{\ii}[1]{\ensuremath{\textcolor{teal}{_{\uparrow#1}}}}
\newcommand{\dd}[1]{\ensuremath{\textcolor{purple}{_{\downarrow#1}}}}

\title{Improving Machine Translation with Human Feedback:\\An Exploration of Quality Estimation as a Reward Model}

\author{%
    Zhiwei He$^1$\thanks{\ \ Work was done when Zhiwei He were interning at Tencent AI Lab.} \quad Xing Wang$^2$\footnotemark[2] \quad Wenxiang Jiao$^2$\quad Zhuosheng Zhang$^1$\\
    \bf Rui Wang$^1$\Thanks{\ \ Xing Wang and Rui Wang are co‐corresponding authors.}$\,$ \quad Shuming Shi$^2$\quad Zhaopeng Tu$^2$\\
    $^1$Shanghai Jiao Tong University\ \ \ $^2$Tencent AI Lab\\
    \texttt{{\{zwhe.cs,zhangzs,wangrui12\}}@sjtu.edu.cn}\\
    \texttt{{\{brightxwang,joelwxjiao,zptu,shumingshi\}}tencent.com}
}

\begin{document}
\maketitle
\begin{abstract}
Insufficient modeling of human preferences within the reward model is a major obstacle for leveraging human feedback to improve translation quality.
Fortunately,  quality estimation (QE), which predicts the quality of a given translation without reference, has achieved impressive alignment with human evaluations in the last two years.
In this work, we investigate the potential of employing the QE model as the reward model to predict human preferences for feedback training.
We first identify the \textit{overoptimization} problem during QE-based feedback training, manifested as an increase in reward while translation quality declines.
We examine the problem and argue that the vulnerability of the QE model might lead to high rewards for incorrect translations, resulting in overoptimization and error propagation.
To address the problem, we adopt a simple yet effective method that uses heuristic rules to detect the incorrect translations and assigns a penalty term to the reward scores of them.
Experimental results show that the proposed QE-based feedback training achieves consistent and significant improvements across various settings, further verified through human preference studies.
Our subsequent analysis demonstrates the high data efficiency of the proposed QE-based feedback training: it outperforms systems using larger parallel corpora by a small amount of monolingual data.
Our code is available at: \url{https://github.com/zwhe99/FeedbackMT}

\end{abstract}
\section{Introduction}
Human feedback has greatly contributed to recent advances in large language models (LLMs), aligning model behavior with human preferences and thereby enhancing the helpfulness and harmlessness of LLMs~\cite{dong2023raft,yuan2023rrhf,zhao2023calibrating,rafailov2023direct}.
The common practice involves using human evaluation data to train a reward model as a proxy for human preferences, followed by feedback training to fine-tune the LLM and maximize the reward score.

Early efforts in neural machine translation (NMT) also attempted to integrate feedback to improve translation quality.
Most works used similarity scores (such as sentence-level BLEU;~\citealt{papineni-etal-2002-bleu}) between the predicted translation and a reference translation to simulate feedback rather than employing feedback from humans~\cite{sokolov-etal-2016-learning,NIPS2016_795c7a7a,kreutzer-etal-2017-bandit,sokolov-etal-2017-shared,lawrence-etal-2017-counterfactual,nguyen-etal-2017-reinforcement,wu-etal-2018-study,wieting-etal-2019-beyond}.
Few attempts to use real human feedback have been made, and these efforts either used implicit feedback in limited scenarios (e.g., e-commerce;~\citealt{kreutzer-etal-2018-neural}) or relied only on a minimal amount of human feedback data~\cite{kreutzer-etal-2018-reliability}.
Therefore, the integration of real human feedback in NMT has been constrained by inadequate modeling of human preferences.

Fortunately, the field of MT has seen substantial advancements in quality estimation (QE;~\citealt{rei-etal-2021-references,rei-etal-2022-cometkiwi,wan-etal-2022-unite}).
A QE model offers a reference-free estimation of translation quality and has been facilitated by the growing availability of human evaluation data~\cite{specia-etal-2020-findings-wmt,specia-etal-2021-findings,zerva-etal-2022-findings} and the development of pre-trained language models~\cite{devlin-etal-2019-bert,conneau-etal-2020-unsupervised}.
Typically, given a source sentence and its translation, a sentence-level QE model can provide a numerical score to indicate the quality of the translation.
The most advanced QE models to date have achieved impressive alignment with human evaluations~\cite{freitag-etal-2022-results} and have been used to guide the decoding process~\cite{fernandes-etal-2022-quality,10.1162/tacl_a_00642}.

In light of this progress, we explore the potential of utilizing QE models as proxies of human preferences and functioning them as reward models in feedback training for the first time.
Firstly, we identify the \textit{overoptimization} problem in feedback training, manifested as an increase in reward while translation quality declines.
Our analysis reveals that the underlying issue lies in the vulnerability of QE-based reward models, which, in rare instances, assign high scores to patently incorrect translations.
As a result, these flawed patterns spread through subsequent training, leading to divergence from human preferences and a training collapse.
However, the reward keeps rising throughout the whole process.
This phenomenon aligns with the observation from \citet{fernandes2023bridging} that ``overoptimizing'' against an imperfect reward model can lead to systems ``that receive good feedback from the model, but not humans''.

Through manual examination of these flawed patterns, we categorize the most prevalent errors into two groups: length ratio errors and off-target errors (i.e., not the desired target language).
Guided by this observation, we propose a simple yet effective solution to mitigate overoptimization.
We first detect these errors and then add a negative penalty to the reward for these erroneous translations.
We show that this approach significantly alleviates the overoptimization problem and results in notable improvements across various settings, which are further verified by human preference studies\footnote{Automatic metrics and human judgement might be not consistent under feedback training as discussed in Limitations.}

In summary, the contributions of this work are detailed as follows:

$\bullet$ We identify the \textit{overoptimization} problem when using QE-based reward models for feedback training and verify the ubiquity of this phenomenon across comprehensive settings (12 in total): 3 QE models $\times$ 2 model architectures (decoder-only LLM and encoder-decoder NMT models) $\times$ 2 resource settings (high-resource and low-resource).
    
$\bullet$ By addressing the overoptimization with a simple yet effective method, we successfully integrate the QE model into feedback training for the first time, achieving remarkable improvements across various settings.
    
$\bullet$ Through further analysis, we demonstrate the high data efficiency of using the QE-based reward model for feedback training, showing that it can outperform systems using larger parallel corpora by only a small amount of monolingual data.

$\bullet$ We investigate the influence of the base model on feedback training, finding that stronger base models (larger in size and pretrained) yield greater improvements after feedback training.
We also examine the effect of crucial hyperparameters.

\section{Feedback Training}
\subsection{Formulation}
We denote an MT model as $M=P(y | x; \theta)$, with model parameters $\theta$.
It takes a source sentence $x$ (or a prompt) as input and generates a target sentence $y$ according to the distribution $P_T(y | x; \theta)$, where $T$ is a temperature parameter to control the diversity.
We consider the QE-based reward model as $r(x,y)$ whose value indicates the quality of $y$ as the translation of $x$.
Taken $\mathcal{D}$ as the training distribution of $x$, the optimization objective is:
\begin{equation}
    \max_{\theta} \mathbb{E}_{x \sim \mathcal{D}, y \sim P(y | x; \theta)} r(x, y).
\end{equation}

We adopt the local ranking version of Reward rAnked FineTuning (RAFT;~\citealt{dong2023raft}) to train the model $M$.
The basic idea of RAFT is to rate the generated candidates from a prompt using the reward model and just learn from the best one of them.
It has been proved to be more stable and efficient than Proximal Policy Optimization (PPO;~\citealt{schulman2017proximal}).
Algorithm \ref{alg:raft} shows the details of RAFT.
\begin{algorithm}[htpb]
    \caption{RAFT}\label{alg:raft}
    \begin{algorithmic}[1]
        \Require Training set $\mathcal{X}$, reward function $r(x, y)$, initial model $M_0=P(y | x; \theta_0)$, batch size $b$, temperature $T$, the number of candidate $k$
        \For{iteration $i$ in ${0,1,\dots, N-1}$}
            \State $D_i \gets \text{SampleBatch}(\mathcal{X}, b)$
            \State $\mathcal{B}=\emptyset$
            \For{$x \in D_i$}
                \State $y_1,\dots,y_k \sim P_T(y|x;\theta_i)$
                \State $y^* = \argmax_{y_j \in \{y_1,\dots,y_k\}} r(x, y_j)$
                \State $\mathcal{B}=\mathcal{B}\cup\{(x, y^*)\}$
            \EndFor
            \State Fine-tune $\theta_i$ on $\mathcal{B}$ to obtain $M_{i+1}=P(y | x; \theta_{i+1})$.
        \EndFor    
    \end{algorithmic}
\end{algorithm}

\subsection{Addressing Overoptimization Problem}
\begin{figure}[htpb]
    \centering
    \includegraphics[width=0.9\linewidth]{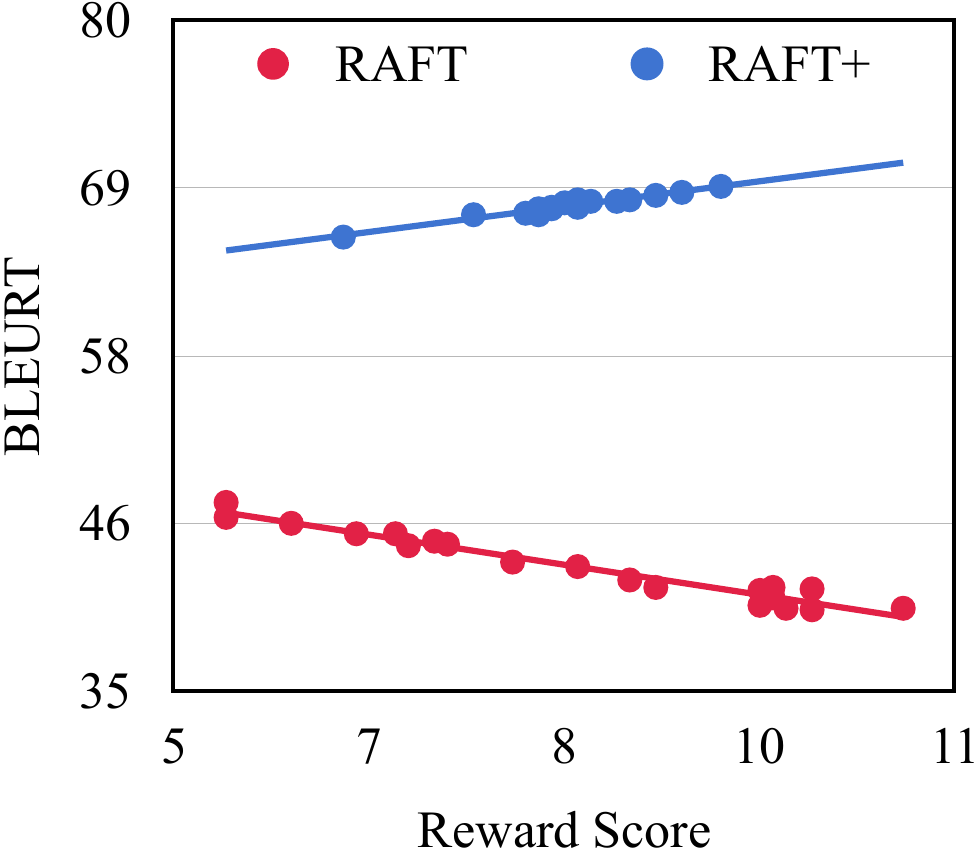}
    \caption{The relationship between reward score and translation quality. Each point represents the average performance of a checkpoint on the development sets. \textsc{Comet-qe-da} is used as QE-based reward model. BLEURT is a translation quality metric that strongly correlates with human preferences~\cite{freitag-etal-2022-results}.}
    \label{fig:overopti-intro}
\end{figure}

\paragraph{Overoptimization} Our preliminary experiment observed \textit{that as the reward increases, the translation performance deteriorates} (shown in \figurename~\ref{fig:overopti-intro}).
This phenomenon is dubbed as \textit{overoptimization} by \citet{gao2023scaling}.
The underlying reason for overoptimization is that the reward model does not serve as a perfect proxy for human preferences.
Hence, overoptimizing rewards could steer the model's behavior away from human preferences.
This aligns with Goodhart's Law, which states, ``When a measure becomes a target, it ceases to be a good measure.''

\begin{figure}[htpb]
    \centering
    \includegraphics[width=0.9\linewidth]{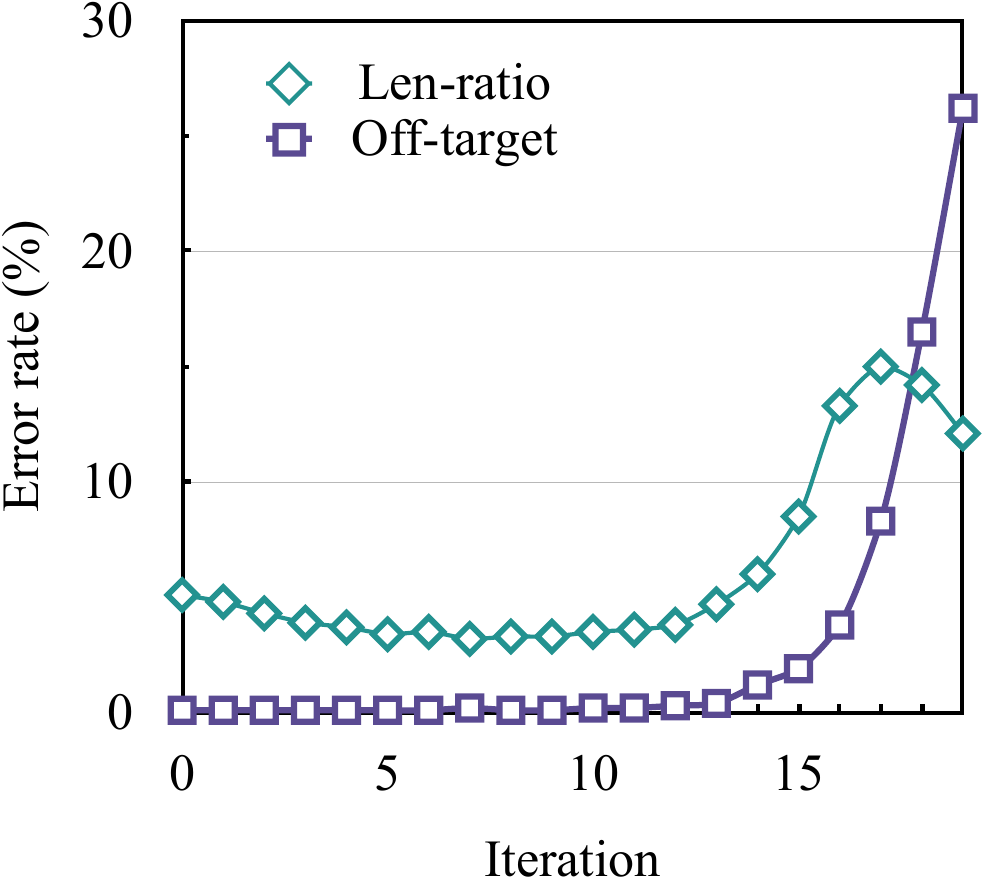}
    \caption{Trends in length-ratio and off-target error rates with the training process. Even though these errors do not manifest significantly during the early and middle phases of training, they may still surge in later stages.}
    \label{fig:diffusion-error}
\end{figure}

\paragraph{Causes}
We further found that the reward model may assign high scores to erroneous translations in some cases (as shown in \tablename~\ref{tab:error-cases}).
These errors encompass common error patterns in MT, such as length-ratio, off-target errors, and hallucinations\footnote{For simplicity, we categorize repeating errors as length-ratio errors, and consider copying-source errors under the category of off-target errors.}.
\begin{table}[htpb]
    \centering
    \resizebox{\linewidth}{!}{
    \begin{tabular}{c|p{4cm}|c}
    \hline
    \bf Error type & \bf Translation & \bf Reward \\
    \hline
    None & The rule of drinking Red Label Whisky:     &  2.84\\
    \hline
    \makecell{Len-ratio \\ (too long/short translation)} & The rule of drinking Red Label Whisky: \hl{1. Always drink responsibly.2. Never drink alone.3. Avoid drinking on an empty stomach.4. Set limits and stick to them.5. Drink in moderation.} & 5.60\\
    \hline
    \makecell{Off-target \\ (wrong target language)} &  \hl{So trinkt man Red-Label-Whisky:} & 4.58\\
    \hline
    \end{tabular}
    }
    \caption{A case of Chinese$\Rightarrow$English translation where the QE model (\textsc{Comet-qe-da}) assigns higher scores to length-ratio and off-target errors than an error-free translation. Error spans are highlighted.}
    \label{tab:error-cases}
\end{table}
While these errors may not be severe initially, they can rapidly propagate to subsequent training stages (see \figurename~\ref{fig:diffusion-error}) once they are accorded high reward scores, which can lead to disruption of the entire training process.
\citet{yan-etal-2023-bleurt} also observed a similar phenomenon when optimizing BLEURT.

\paragraph{Solution}
To alleviate overoptimization, we monitor length-ratio and off-target errors during training and assign negative, punitive rewards for these errors.
Let $C(x,y)$ be true if $\frac{|y|}{|x|} \notin [L, U]$ or $\text{lang}(y) \neq \text{target language}$, where $[L, U]$ is an acceptable length ratio interval and $\text{lang}(\cdot)$ is a language identification function (detailed in~\cref{sec:appendix-error-detection}).
Then the reward modification can be expressed as:
\begin{equation}
r^+(x,y) =
\begin{cases}
r(x,y) - P & \text{if } C(x, y) \\
r(x,y) & \text{otherwise,}
\end{cases}
\end{equation}
where $P$ is the penalty term that we simply set to $-\infty$ since RAFT is an algorithm based on data selection.
We refer to this approach as RAFT+.
As depicted in \figurename~\ref{fig:overopti-intro}, RAFT+ facilitates a trend of concurrent improvement in reward and translation quality.
Notably, the relationship between reward and translation quality approximates a linear correlation.
This suggests that optimizing QE-based reward can be an effective strategy in instances devoid of overoptimization.
\section{Experiments}
\subsection{Experimental Setup}
\paragraph{Pipeline}
We adopt the pipeline of reinforcement learning with human feedback (RLHF;~\citealt{ouyang2022training}).
Starting with the pretrained base model, we carry out the following steps: (1) Supervised fine-tuning (SFT), where we utilize parallel data to fine-tune the pretrained base model, thereby obtaining an initial MT model.
(2) Feedback training, where we train the model using RAFT/RAFT+ to maximize the reward from a QE model.
Note that this stage only uses monolingual data.

\begin{figure*}[t!]
    \centering
    \includegraphics[width=0.9\linewidth]{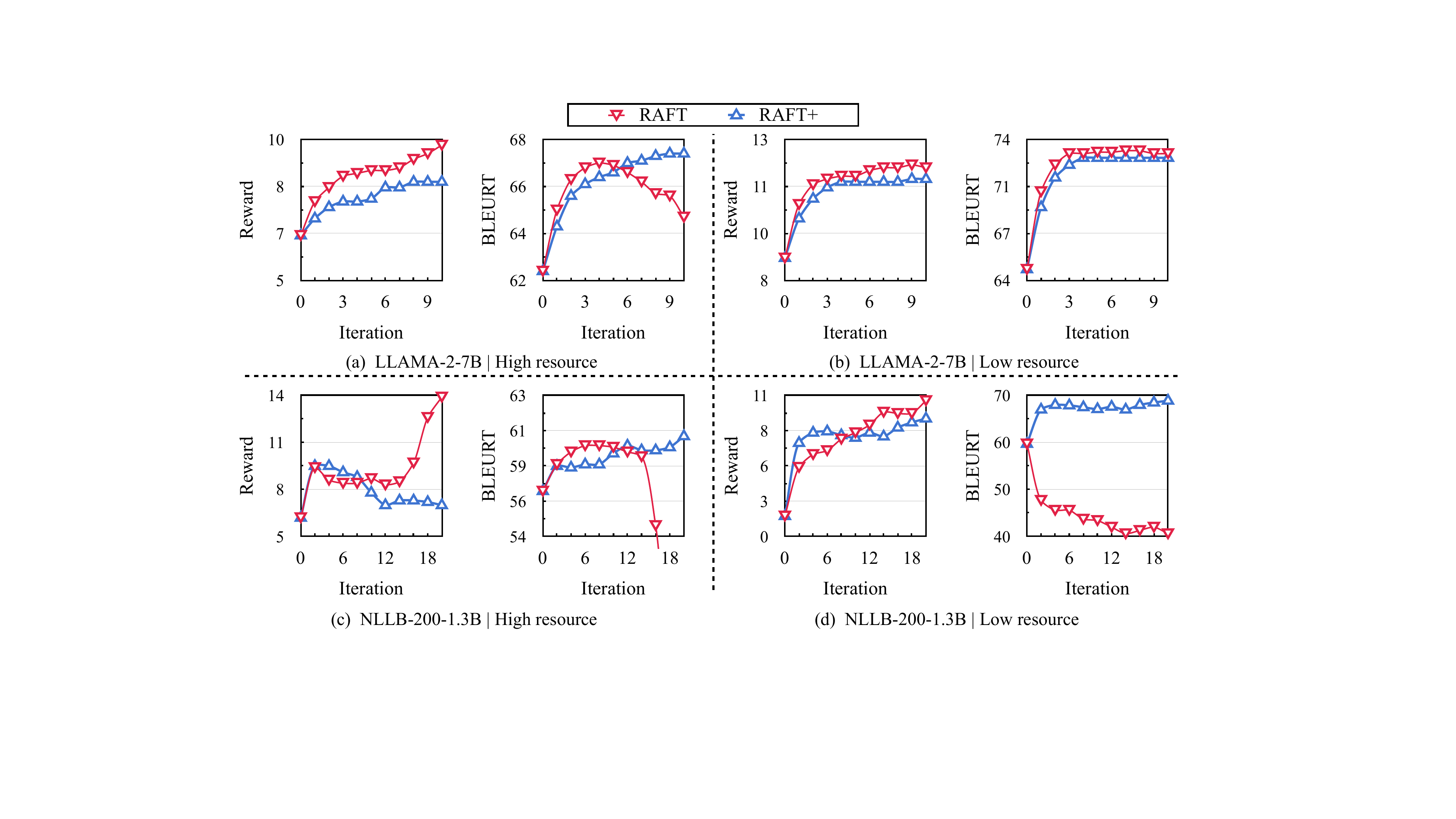}
    \caption{Training curves under various settings. The metrics are average values for all language pairs on the development set. The QE-based reward model is \textsc{Comet-qe-da}.}
    \label{fig:training-curve}
\end{figure*}

\paragraph{Model}
For the base models, we adopt:
\begin{itemize}[leftmargin=10pt]
    \item \textsc{Llama-2-7B}~\cite{touvron2023llama}: a decoder-only LLM primarily trained in English, with the objective to predict the next token.
    \item \textsc{Nllb-200-1.3B}~\cite{costa2022no}: an encoder-decoder model that trained for multilingual translation across 200 languages.
    After SFT, it will be adapted to the language pairs we considered.
\end{itemize}
For QE models, we use:
\begin{itemize}[leftmargin=10pt]
    \item \textsc{Comet-qe-da}~\cite{rei-etal-2021-references}: A QE model trained on \textit{Direct Assessments} data.
    \item \textsc{Comet-qe-mqm}~\cite{rei-etal-2021-references}: Fine-tuned \textsc{Comet-qe-da} model using \textit{Multidimensional Quality Metric} (MQM) data.
    \item \textsc{UniTE-MUP}~\cite{wan-etal-2022-unite}: A unified translation evaluation framework that is jointly trained for reference-only, source-only, and source-reference-combined evaluation.
    We only use its source-only evaluation.
\end{itemize}

\paragraph{Data}
We consider the settings for both high- and low-resource language pairs.
In the high-resource setting, we focus on English$\Leftrightarrow$Chinese and English$\Leftrightarrow$German.
In the low-resource setting, our focus is on English$\Leftrightarrow$Ukrainian and Ukrainian$\Leftrightarrow$Czech.
It is important to note that both settings are multilingual, where all four translation directions are trained concurrently within one model, and we ensure the balance of all directions to avoid introducing other factors.
That is, every direction has an equal number of training samples.
We adopt WMT22~\cite{kocmi-etal-2022-findings,he-etal-2022-tencent} as the test sets for all language pairs, which is latest enough to avoid data contamination~\cite{zhu2023clean,zhu2023multilingual}.
For development sets, high resource setting uses WMT21, and low resource setting uses Flores200~\cite{goyal-etal-2022-flores}.
\tablename~\ref{tab:data-stat} lists the detail of the data we used.
\begin{table}[htpb]
    \centering
    \resizebox{0.7\linewidth}{!}{
    \begin{tabular}{c l l l}
    \toprule
    & & \bf High-res & \bf Low-res \\
    \midrule
    \multirow{2}{*}[-10pt]{\bf Train} & SFT (para)  & WMT (2M) & Wiki (200K)\\
    \cmidrule{2-4}
                                      & FB (mono)   & CC (85K) & Wiki (84K) \\
    \midrule
    \multicolumn{2}{c}{\bf Dev}       &  WMT21 & Flores200\\
    \multicolumn{2}{c}{\bf Test}      &  WMT22 & WMT22\\
    \bottomrule
    \end{tabular}}
    \caption{Data used in the different phases of our pipeline. SFT uses parallel data, while FB uses monolingual data. Numbers indicate the number of samples. ``FB'' denotes feedback training. ``CC'' and ``Wiki'' denote CCMatrix~\cite{schwenk-etal-2021-ccmatrix}\protect\footnotemark and WikiMatrix~\cite{schwenk-etal-2021-wikimatrix}\protect\footnotemark. WMT and CC are both general domain.}
    \label{tab:data-stat}
\end{table}
\footnotetext{\url{https://github.com/facebookresearch/LASER/tree/main/tasks/CCMatrix}}
\footnotetext{\url{https://github.com/facebookresearch/LASER/tree/main/tasks/WikiMatrix}}

\paragraph{Training details}
For SFT, we conducted training for one epoch.
For RAFT/RAFT+, we trained \textsc{Llama-2-7B} for 10 iterations using a learning rate of 2e-6, and for \textsc{Nllb-200-1.3B}, we trained for 20 iterations with a learning rate of 2e-5.
We set the batch size $b$ to 1024, the number of candidates $k$ to 8, and the temperature $T$ to 0.85.

\paragraph{Evaluation}
We use \textsc{COMET}~\cite{rei-etal-2022-comet} and \textsc{BLEURT}~\cite{sellam-etal-2020-bleurt} to assess translation quality.
These neural metrics show superiority over string-based metrics like BLEU~\cite{freitag-etal-2022-results,kocmi-etal-2021-ship,bawden2023investigating}.
We use \texttt{Unbabel/wmt22-comet-da} and \texttt{BLEURT-20} checkpoints for these two metrics.
It is worth noting that we also observe the overoptimization problem on COMET, that is, COMET increased but the actual translation quality decreased (see \cref{sec:results}).
Therefore, we use BLEURT as the main metric, but still report COMET as a reference and conduct human evaluation in \cref{sec:human-evaluation}.
We select the best checkpoint based on the performance on the dev sets, and report the final results on test sets using beam search (beam size = 4).

\subsection{Results}
\label{sec:results}
\paragraph{Training curves}
\figurename~\ref{fig:training-curve} illustrates the training curves of reward and BLEURT on the development sets when using \textsc{Comet-qe-da} as the QE-based reward model.
Our observations are:

$\bullet$ \textbf{Overoptimization is a phenomenon of high frequency when using vanilla RAFT.}
Three out of the four settings (\figurename~\ref{fig:training-curve}a, c, d) have the situation that the reward shows an increase while the BLEURT declines.
We further verify this phenomenon in \cref{sec:appendix-more-training-curves} when using \textsc{Comet-qe-mqm} or \textsc{UniTE-MUP} as the reward model.

$\bullet$ \textbf{The severity of overoptimization varies under different settings.}
\figurename~\ref{fig:training-curve}d represents the most severe overoptimization, where the BLEURT score starts decreasing from the onset of the training process.
\figurename~\ref{fig:training-curve}a and \figurename~\ref{fig:training-curve}c exhibit a trend of initial increase followed by a decrease.
In such scenarios, a relatively good checkpoint can be chosen based on the performance on the development set.
Conversely, \figurename~\ref{fig:training-curve}b does not display overoptimization.
We conjecture that the severity of overoptimization could be related to multiple factors, including language pairs, the reward model, and the SFT model.

$\bullet$ \textbf{RAFT+ alleviates overoptimization effectively.}
The BLEURT scores of RAFT+ in the four settings consistently increase as the training progresses.
At the same time, the growth rate of the reward scores in RAFT+ is significantly slower than that in RAFT.
We even witness a situation where the reward decreases in \figurename~\ref{fig:training-curve}c.
Moreover, in \figurename~\ref{fig:training-curve}b where overoptimization did not occur originally, RAFT+ can still achieve a performance close to that of RAFT.
However, it is worth noting that the overoptimization problem has not been completely solved, as some errors, like hallucinations, are challenging to identify.
Furthermore, we cannot guarantee generalization to all other language pairs since language identification is not reliable for extremely low-resource language~\cite{aji-etal-2022-one}, indicating room for improvement.
We leave a more comprehensive study to future work.

\begin{table*}[t!]
    \centering
    \begin{subtable}{\textwidth}
    \centering
    \resizebox{0.9\linewidth}{!}{
    \begin{tabular}{l ccccccccll}
        \toprule
        \multirow{2}{*}{ Method} & \multicolumn{2}{c}{ De$\Rightarrow$En} & \multicolumn{2}{c}{ En$\Rightarrow$De} & \multicolumn{2}{c}{ Zh$\Rightarrow$En} & \multicolumn{2}{c}{ En$\Rightarrow$Zh} & \multicolumn{2}{c}{\bf Average} \\
        \cmidrule(lr){2-3}\cmidrule(lr){4-5}\cmidrule(lr){6-7}\cmidrule(lr){8-9}\cmidrule(lr){10-11}
        & {\small COMET} & {\small BLEURT}   & {\small COMET} & {\small BLEURT} & {\small COMET} & {\small BLEURT} & {\small COMET} & {\small BLEURT} & {\small\bf COMET} & {\small\bf BLEURT} \\
        \midrule
        WMT22 Best       &   85.0 &   73.8 &   87.4 &   77.9 &   81.0 &   68.9 &   86.8 &   72.8 &   85.1            &   73.4\\
        \midrule
        \rowcolor{gray!25}
        \multicolumn{11}{c}{\textsc{ Llama-2-7B}}    \\
        \textsc{SFT}     &   82.5 &   70.5 &   80.7 &   68.2 &   76.1 &   62.3 &   84.9 &   69.3 &   81.0            &   67.6\\
        \midrule
        \multicolumn{11}{l}{\textsc{Reward model: Comet-qe-da}}    \\[3pt]
        \textsc{~~RAFT}  &   83.7 &   72.1 &   82.8 &   71.1 &   78.7 &   65.3 &   85.9 &   70.1 &   82.8\ii{1.7}    &   69.7\ii{2.1}\\
        \textsc{~~RAFT+} &   83.6 &   72.1 &   84.4 &   73.9 &   79.0 &   66.1 &   85.4 &   69.3 &\bf83.1\ii{\bf2.1} &\bf70.3\ii{\bf2.7}\\
        \midrule
        \multicolumn{11}{l}{\textsc{Reward model: Comet-qe-mqm}}    \\[3pt]
        \textsc{~~RAFT}  &   83.3 &   72.0 &   84.8 &   75.1 &   77.8 &   64.3 &   86.1 &   70.4 &   83.0\ii{2.0}    &   70.5\ii{2.9}\\
        \textsc{~~RAFT+} &   83.7 &   72.4 &   85.6 &   75.7 &   78.6 &   65.6 &   85.8 &   70.0 &\bf83.4\ii{\bf2.4} &\bf70.9\ii{\bf3.3}\\
        \midrule
        \midrule
        \rowcolor{gray!25}
        \multicolumn{11}{c}{\textsc{ Nllb-200-1.3B}}    \\
        \textsc{SFT}     &   70.9 &   52.5 &   85.3 &   74.8 &   66.0 &   48.4 &   83.7 &   69.1 &   76.5            &   61.2\\
        \midrule
        \multicolumn{11}{l}{\textsc{Reward model: Comet-qe-da}}    \\[3pt]
        \textsc{~~RAFT}  &   73.2 &   52.2 &   85.8 &   75.1 &   67.9 &   50.5 &   84.2 &   68.9 &   77.8\ii{1.3}    &   61.7\ii{0.5}\\
        \textsc{~~RAFT+} &   74.2 &   56.7 &   85.8 &   75.2 &   69.0 &   52.6 &   84.0 &   67.9 &\bf78.2\ii{\bf1.7} &\bf63.1\ii{\bf1.9}\\
        \midrule
        \multicolumn{11}{l}{\textsc{Reward model: Comet-qe-mqm}}    \\[3pt]
        \textsc{~~RAFT}  &   82.8 &   71.3 &   83.9 &   73.4 &   76.1 &   62.3 &   84.6 &   68.6 &   81.8\ii{5.3}    &   68.9\ii{7.7}\\
        \textsc{~~RAFT+} &   83.3 &   71.8 &   84.6 &   74.4 &   76.7 &   62.9 &   84.6 &   68.4 &\bf82.3\ii{\bf5.8} &\bf69.4\ii{\bf8.2}\\
        \bottomrule
    \end{tabular}}
    \caption{High-resource language pairs}
    \label{tab:main-results-high}
    \end{subtable}

    \vspace{10pt}

    \begin{subtable}{\textwidth}
    \centering
    \resizebox{0.9\linewidth}{!}{
    \begin{tabular}{l ccccccccll}
        \toprule
        \multirow{2}{*}{ Method} & \multicolumn{2}{c}{ En$\Rightarrow$Uk} & \multicolumn{2}{c}{ Uk$\Rightarrow$En} & \multicolumn{2}{c}{ Uk$\Rightarrow$Cs} & \multicolumn{2}{c}{ Cs$\Rightarrow$Uk} & \multicolumn{2}{c}{\bf Average} \\
        \cmidrule(lr){2-3}\cmidrule(lr){4-5}\cmidrule(lr){6-7}\cmidrule(lr){8-9}\cmidrule(lr){10-11}
        & {\small COMET} & {\small BLEURT}   & {\small COMET} & {\small BLEURT} & {\small COMET} & {\small BLEURT} & {\small COMET} & {\small BLEURT} & {\small\bf COMET} & {\small\bf BLEURT} \\
        \midrule
        WMT22 Best       &   87.8 &   76.5 &   85.9 &   76.6 &   92.2 &   82.8 &   91.6 &   80.3 &   89.4            &   79.1\\
        \midrule
        \rowcolor{gray!25}
        \multicolumn{11}{c}{\textsc{ Llama-2-7B}}    \\
        \textsc{SFT}     &   79.2 &   64.0 &   76.7 &   66.0 &   70.0 &   53.2 &   71.2 &   51.3 &   74.3            &   58.6\\
        \midrule
        \multicolumn{11}{l}{\textsc{Reward model: Comet-qe-da}}    \\[3pt]
        \textsc{~~RAFT}  &   82.3 &   68.0 &   81.4 &   71.1 &   82.5 &   69.5 &   84.3 &   69.9 &\bf82.6\ii{\bf8.3} &\bf69.6\ii{\bf11.0}\\
        \textsc{~~RAFT+} &   82.0 &   67.8 &   81.5 &   71.2 &   82.2 &   68.8 &   84.5 &   70.1 &\bf82.6\ii{\bf8.3} &   69.5\ii{10.9}\\
        \midrule
        \multicolumn{11}{l}{\textsc{Reward model: Comet-qe-mqm}}    \\[3pt]
        \textsc{~~RAFT}  &   80.7 &   65.5 &   76.7 &   66.0 &   75.7 &   59.9 &   75.2 &   54.8 &   77.1\ii{2.8}    &   61.5\ii{2.9}\\
        \textsc{~~RAFT+} &   81.2 &   67.0 &   79.2 &   68.9 &   77.3 &   62.3 &   78.8 &   60.7 &\bf79.1\ii{\bf4.8} &\bf64.8\ii{\bf6.2}\\
        \midrule
        \midrule
        \rowcolor{gray!25}
        \multicolumn{11}{c}{\textsc{ Nllb-200-1.3B}}    \\
        \textsc{SFT}     &   83.1 &   70.2 &   71.1 &   62.7 &   73.2 &   61.5 &   57.3 &   43.4 &   71.2            &   59.4\\
        \midrule
        \multicolumn{11}{l}{\textsc{Reward model: Comet-qe-da}}    \\[3pt]
        \textsc{~~RAFT}  &   85.2 &   72.5 &   64.7 &   33.2 &   70.5 &   29.7 &   73.8 &   30.1 &   73.6\ii{2.4}    &   41.4\dd{18.0}\\
        \textsc~~{RAFT+} &   84.5 &   71.3 &   77.7 &   67.0 &   83.1 &   70.3 &   72.0 &   55.1 &\bf79.3\ii{\bf8.1} &\bf65.9\ii{\bf6.6}\\
        \midrule
        \multicolumn{11}{l}{\textsc{Reward model: Comet-qe-mqm}}    \\[3pt]
        \textsc{~~RAFT}  &   85.8 &   73.2 &   67.5 &   50.0 &   71.1 &   41.6 &   71.1 &   42.7 &   73.9\ii{2.7}    &   51.9\dd{7.5}\\
        \textsc{~~RAFT+} &   84.5 &   71.8 &   76.4 &   66.1 &   82.1 &   69.9 &   71.4 &   54.5 &\bf78.6\ii{\bf7.4} &\bf65.6\ii{\bf6.2}\\
        \bottomrule
    \end{tabular}}
    \caption{Low-resource language pairs}
    \label{tab:main-results-low}
    \end{subtable}
    \caption{Translation performance on the test sets under various settings, using \textsc{Comet-qe-da} and \textsc{Comet-qe-mqm} as reward models. Results when \textsc{UniTE-MUP} is used as the reward model are presented in \cref{sec:appendix-unite-mup-as-the-reward-model}. Bold indicates that the average performance of the method exceeds that of SFT and RAFT/RAFT+ within the same QE model. The subscripts indicate the change in performance relative to the SFT. ``WMT22 Best'' means competition winners for each direction as reported in WMT22 correspond to those used by~\citet{gpt-mt-2023}.}
    \label{tab:main-results}
\end{table*}

\paragraph{Main results}
\tablename~\ref{tab:main-results} shows the main results on test sets when using \textsc{Comet-qe-da} or \textsc{Comet-qe-mqm} as the QE-based reward model.
We present the results of \textsc{UniTE-MUP} in \cref{sec:appendix-unite-mup-as-the-reward-model} and use chrF as the evaluation metric in \cref{sec:appendix-chrf-as-the-evaluation-metric}.

$\bullet$ \textbf{Feedback training brings significant improvements in general.}
Regardless of whether the resource setting is high or low and irrespective of the variation in base models, RAFT+ tends to deliver notable enhancements (though still under-performs the WMT22 Best systems), particularly pronounced in the low resource setting.
This suggests that current QE models are already equipped with the capability to function as reward models after addressing the overoptimization problem.
In contrast, while achieving good gains in most settings, RAFT can suffer from severe overoptimization that might lead to failed training.

$\bullet$ \textbf{The performance of the reward model varies under different resource settings.}
In the high-resource setting, \textsc{Comet-qe-mqm} outperforms \textsc{Comet-qe-da}, while the opposite is true in the low-resource setting, which suggests that QE models still have a lot of room for improvement.

$\bullet$ \textbf{Remarkably, even COMET, a reference-based metric, can be overoptimized.}
In RAFT training of \textsc{Nllb-200-1.3B} under low-resource setting (bottom of \tablename~\ref{tab:main-results-low}), there is a marginal increase in COMET scores, yet a significant drop in BLEURT, which is unusual.
By inspecting the outputs, we found that severe off-target errors significantly hamper the model's performance.
One plausible reason might be the similarities between the COMET and COMET-QE models, both of which may be susceptible to these off-target errors.
This finding underscores the necessity of treating automatic metrics with caution and the importance of employing multiple metrics simultaneously for a more comprehensive evaluation.
Especially for feedback training using a reward model, the increasement of automatic metrics might not indicate true translation quality improvement.
We therefore conducted a human evaluation in \cref{sec:human-evaluation} and have a discussion in Limitations.
It also stresses the significance of recognizing a metric's vulnerabilities during its development, including adjustments for prevalent translation inaccuracies such as length-ratio, off-target errors, hallucinations.
\section{Analysis}
\subsection{Human evaluation}
\label{sec:human-evaluation}
\begin{figure}[htpb]
    \centering
    \includegraphics[width=\linewidth]{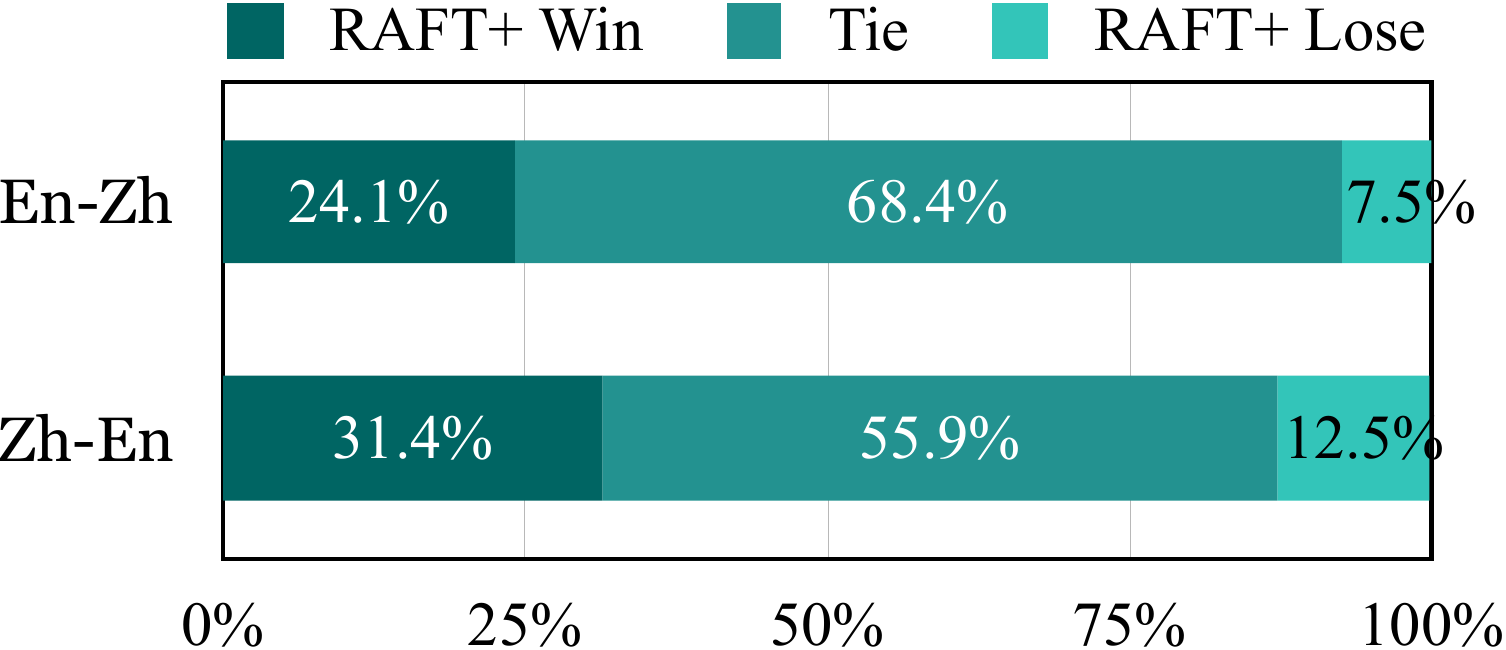}
    \caption{Human preference evaluation, comparing RAFT+ to SFT model on En$\Leftrightarrow$Zh test sets.}
    \label{fig:human-preference}
\end{figure}
As discussed in \cref{sec:results}, automatic metrics may not correlate perfectly with actual translation quality.
Therefore, we perform human preference studies on En$\Leftrightarrow$Zh test sets.
For each test sample, our annotators (professional translators) were presented with a source sentence and two possible translations generated by either the SFT or RAFT+ model (based on \textsc{Llama-2-7B}).
They were then tasked with selecting the superior translation or determining that neither translation was better than the other.
\figurename~\ref{fig:human-preference} shows the results of human preference studies.
We find that RAFT+ achieves better or equal translations in 92.5\% of the cases for En-Zh and 87.3\% for Zh-En, compared to SFT, confirming the effectiveness of feedback training.

\subsection{Data Efficiency of Feedback Training}
\label{sec:data-efficiency-of-feedback-training}
\begin{figure}[htpb]
    \centering
    \includegraphics[width=0.8\linewidth]{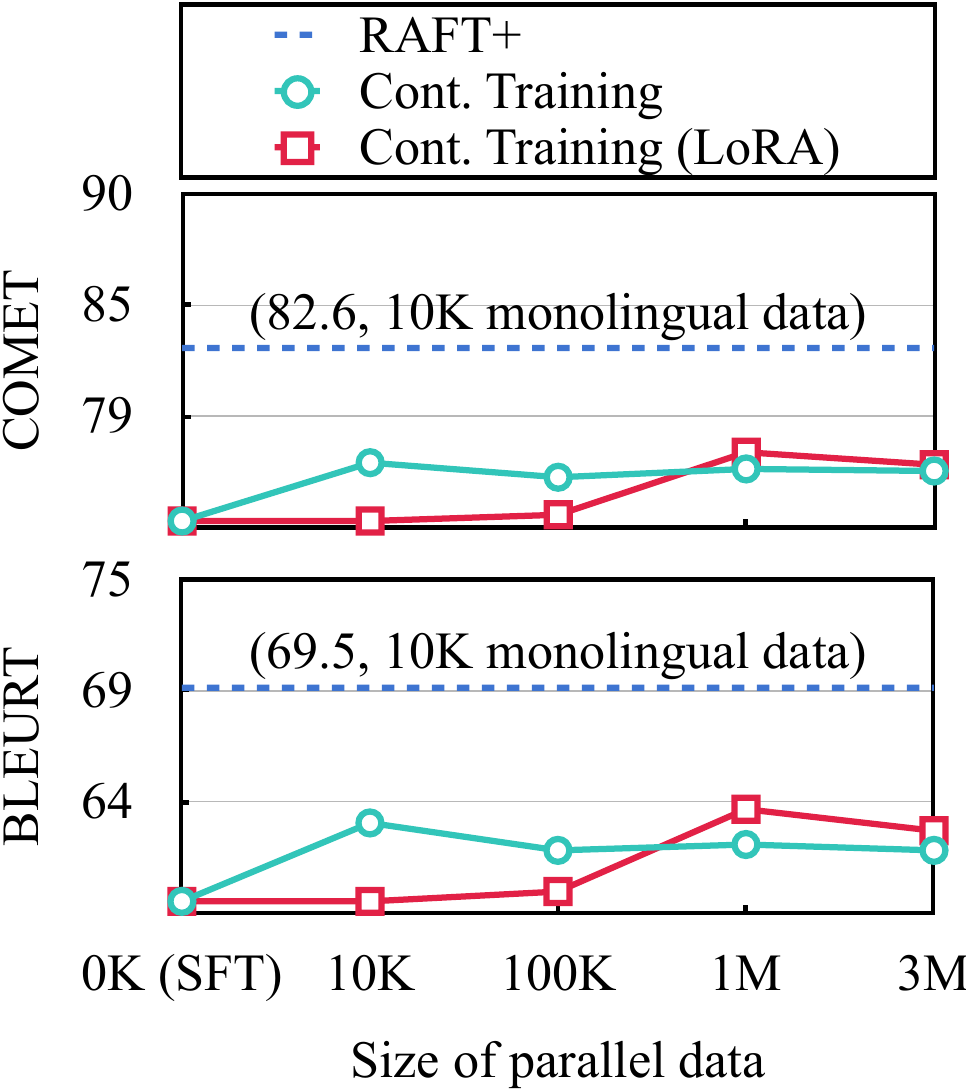}
    \caption{Comparison between RAFT+ and continuous training in the low-resource setting.}
    \label{fig:data-efficiency}
\end{figure}
Data efficiency refers to the capacity to achieve good performance with less training data.
For low-resource language pairs, data efficiency is important since it is costly to annotate high-quality parallel data~\cite{he-etal-2022-bridging}.
Feedback training provides an alternative path where humans only need to evaluate rather than generate translations.

We examine the data efficiency of feedback training in the low-resource setting.
We collected an extra 3M of parallel data from WikiMatrix~\cite{schwenk-etal-2021-wikimatrix} and continued training the SFT model (based on $\textsc{Llama-2-7B}$) using data of different sizes.
We adopted two continued training strategies: full parameter fine-tuning and parameter-efficient fine-tuning, specifically employing LoRA~\cite{hu2022lora}.
To be fair, we filtered samples with length-ratio and off-target errors in the parallel data.
On the other hand, RAFT+ only consumes 10K monolingual data (1024 batch size $\times$ 10 iterations).
\figurename~\ref{fig:data-efficiency} depicts their average performance on test sets.
Unexpectedly, the continuous training with increasing amounts of parallel data fails to yield consistent improvements.
This observation aligns with a similar phenomenon reported by~\citet{zhou2023lima}.
A plausible explanation could be the low quality of the crawled data for low-resource languages.
Conversely, RAFT+ performs markedly better using merely 10K monolingual data, exhibiting high data efficiency.
It is also worth noting that in RAFT+ the model is only trained on self-generated translations.
Therefore, we conjecture that the SFT model already has strong translation potential inherently, and RAFT+ can bring this potential to fruition.

\subsection{Effects of Scaling Model Size and Pretraining}
\begin{figure}[htpb]
    \centering
    \includegraphics[width=0.83\linewidth]{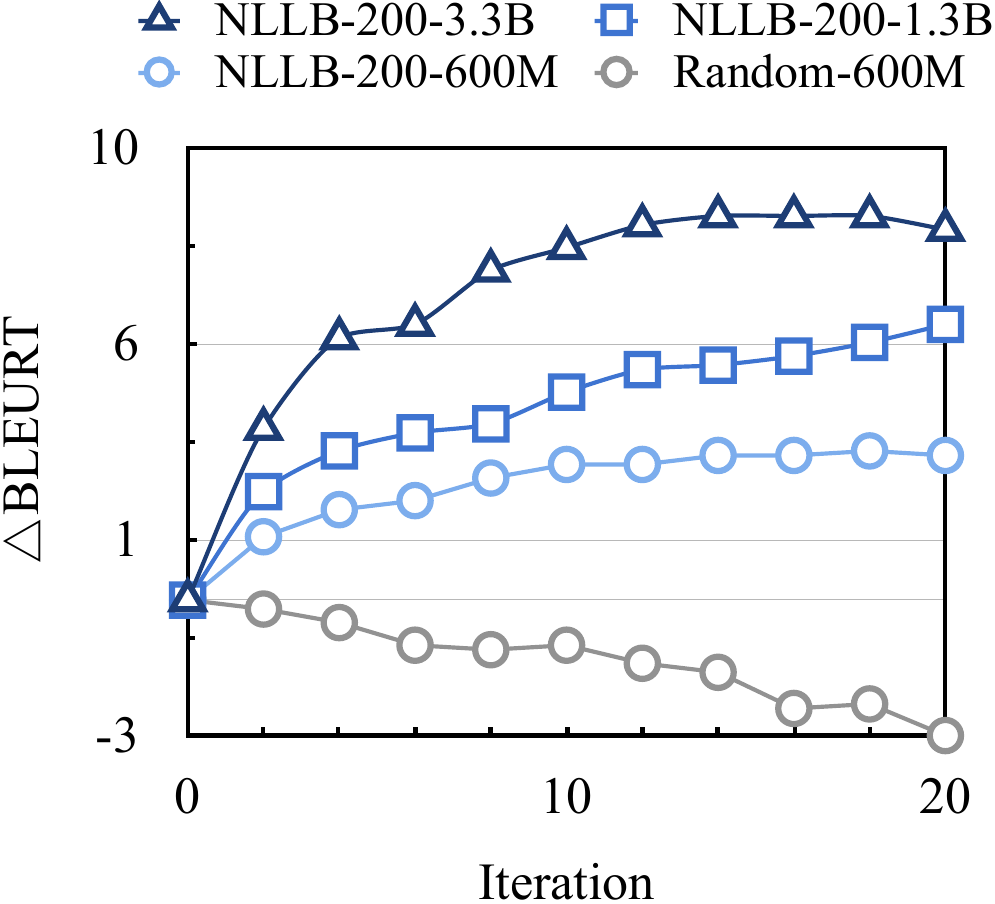}
    \caption{Training curves of RAFT+ (high-resource, \textsc{Comet-qe-mqm}) under different base models. We report the change in BLEURT score for each checkpoint relative to the SFT model.}
    \label{fig:pretrain-data-and-scaling-model-size}
\end{figure}

In this section, we examine the effects of two dimensions of the base model on feedback training: model size and the presence of pretraining.
We conduct experiments under the high-resource setting, using \textsc{Comet-qe-mqm} as the reward model.
For model size, we consider NLLB-200-(600M, 1.3B, 3.3B) as base models; for pretraining, we randomly initialize NLLB-200-600M and perform the SFT from scratch with 80M parallel data, which we call Random-600M.
From \figurename~\ref{fig:pretrain-data-and-scaling-model-size}, we have two obvious phenomena: (1) a larger base model size results in a more significant enhancement from feedback training; (2) feedback training is effective only when the base model has undergone pretraining.
Combining the two, we deduce that a stronger base model results in more significant improvements from feedback training.
A plausible rationale for this could be that a well-established base model inherently possesses great potential, which can be further unlocked by suitable feedback.

\subsection{Effects of Hyperparameters}
\begin{figure}[htpb]
    \centering
    \includegraphics[width=\linewidth]{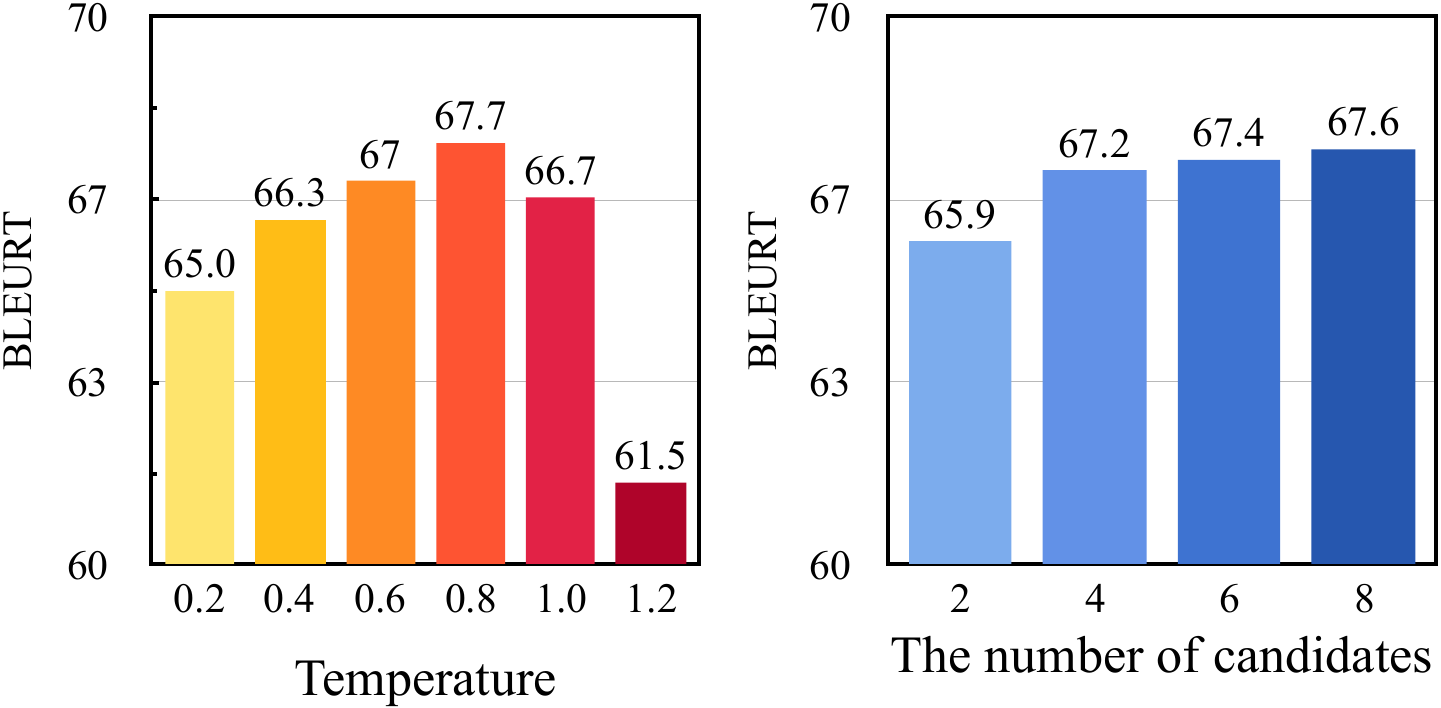}
    \caption{Effects of temperature $T$ and the number of candidates $k$ on the final performance (high-resource setting, \textsc{Llama-2-7B}, \textsc{Comet-qe-da}).}
    \label{fig:hyperpara}
\end{figure}
The sampling temperature and the number of candidates are two crucial hyperparameters for RAFT+.
Both influence the diversity of the candidates.
Intuitively, having more diverse candidates may provide richer feedback signals, potentially leading to better final performance.
However, as illustrated in \figurename~\ref{fig:hyperpara}, (1) continuously increasing the temperature leads to performance degradation; (2) the boost from continuously increasing the number of candidates approaches saturation.
This means that the increase in candidate diversity has reached its upper limit.
\section{Related Work}
\subsection{Feedback Training in MT}
Many early works have attempted to use various forms of feedback to improve MT, where the model has no access to the gold references but receives partial feedback on its output.
Depending on the source of the feedback, they can be categorized into simulated or human feedback.

\paragraph{Simulated feedback}
This type of work utilizes a measure of similarity between the model's output and the reference translation to simulate feedback, e.g., sentence-BLEU and cross-entropy loss~\cite{sokolov-etal-2016-learning,NIPS2016_795c7a7a}.
\citet{kreutzer-etal-2017-bandit} lifts linear bandit learning to neural sequence-to-sequence learning.
\citet{lawrence-etal-2017-counterfactual} demonstrates the possibility of counterfactual learning from deterministic bandit logs.
\citet{nguyen-etal-2017-reinforcement} propose a training algorithm combining the advantage actor-critic algorithm with the attention-based neural encoder-decoder architecture.
\citet{wieting-etal-2019-beyond} proposes to use semantic 
 similarity rather than BLEU as simulated feedback.
\citet{wu-etal-2018-study} discusses how to train the MT model using feedback effectively.
The main drawback of simulated feedback is that there is no real human involvement, so the metrics being optimized, such as BLEU, do not correlate well with human preferences~\cite{freitag-etal-2022-results}.
In addition, the need for references presents a challenge for low-resource language pairs.

\paragraph{Human feedback}
\citet{kreutzer-etal-2018-neural} utilizes implicit human feedback in a constrained e-commerce scenario, which cannot be applied to general MT.
\citet{kreutzer-etal-2018-reliability} investigates how to improve translation using human evaluation data but is limited by the small data size (1K).
Concurrently with this work, \citet{finkelstein2023mbr,ramos2023aligning} and \citet{gulcehre2023reinforced} also consider using QE models to provide reward signals in the training of MT models.
\citet{xu2024contrastive} points out that the references of training data might be worse than model outputs in quality, which aligns with our findings in \cref{sec:data-efficiency-of-feedback-training}.

\subsection{Aligning LLMs with Human}
Pretrained on raw text from the internet, LLMs might generate toxic, inaccurate, and unhelpful content~\cite{fernandes2023bridging}.
To mitigate these issues, researchers have employed human feedback to better align the behavior of LLMs with human preferences, thereby enhancing their helpfulness and reducing potential harm~\cite{ouyang2022training,stiennon2020learning,bai2022training}.
The basic idea involves learning a reward function that captures human preferences and optimizing the LLM using proximal policy optimization (PPO;~\citealt{schulman2017proximal}) to maximize the reward.
Lighter weight training strategies, such as RAFT~\cite{dong2023raft}, RRHF~\cite{yuan2023rrhf} and DPO~\cite{rafailov2024direct}, use training data ranking based on rewards or human annotations to align the LLM.
LIMA~\cite{zhou2023lima} underscores the importance of high-quality supervised data for effective alignment.
For a comprehensive understanding of alignment, we recommend readers refer to survey papers:~\citet{fernandes2023bridging} and~\citet{aligning_llm_human}.

\section{Conclusion}
We explore the potential of using the current QE model as a reward model.
By identifying, analyzing, and mitigating the overoptimization problem, we successfully integrate the QE model into feedback training to refine translation quality.
We validate its effectiveness across various settings, including human evaluation.
Further analysis demonstrates the high data efficiency of feedback training using the QE-based reward model.
Lastly, we delve into the impact of the base model and hyperparameters on feedback training.

\section*{Acknowledgements}
Zhiwei and Rui are with MT-Lab, Department of Computer Science and Engineering, School of Electronic Information and Electrical Engineering, and also with the MoE Key Lab of Artificial Intelligence, AI Institute, Shanghai Jiao Tong University, Shanghai 200204, China.  Rui and Zhiwei are supported by the Tencent Open Fund (RBFR2023012), the National Natural Science Foundation of China (62176153), and the Shanghai Municipal Science and Technology Major Project (2021SHZDZX0102).

\section*{Limitations}
\label{sec:limitation}

\paragraph{\textcolor{purple}{An increase in automatic metrics do not necessarily indicate a true improvement in translation quality.}}
When we conducting feedback training, we need a QE model to provide reward score and a metric model to evaluate the final performance.
Though they play different roles in the whole pipeline, they share the same function of assessing translation quality.
Unfortunately, current open-sourcing QE and metric models that show strong correlation with human mostly rely on human evaluation data in WMT, which indicating the overlap between their training data.
Therefore, overoptimizing the QE model might lead to unconsciously overoptimizing the metric model, as we observed in \cref{sec:results}.
This is the reason why we conducted a human preference study to double check the improvement.
However, human annotation is always costly and not scalable, and we only conduct it on En$\Leftrightarrow$Zh.

\paragraph{Training Algorithm}
This work only considers a simple yet stable training algorithm, RAFT, in favor of its ease of use and reduced computational requirements.
We also implemented the Minimum Risk Training (MRT) algorithm~\cite{shen-etal-2016-minimum} in \cref{sec:appendix-mrt-implementation}, but found it difficult to train stably.
While other more commonly used reinforcement learning (RL) algorithms such as PPO exist, their training stability can be influenced by a variety of factors and often demands substantial GPU memory~\cite{dong2023raft,zheng2023secrets}.
Hyperparameters, stability, and computational constraints led us to consciously limit our exploration to RAFT, thereby not investigating other potentially effective but more complex training algorithms.
This choice simplifies the training process but might limit the general applicability of the approach in different contexts.
Future work may explore these alternative algorithms, acknowledging the trade-offs between complexity, stability, and performance.

\paragraph{Granularity of Quality Estimation}
This work exclusively focuses on sentence-level QE, neglecting other granularities, such as word-level or document-level QE.
The choice of granularity inherently limits the scope of our insights and applications.
Understanding how word-level QE feedback might be utilized, or how QE could be employed to enhance document translation quality, presents exciting avenues for future research.

\paragraph{Imbalance in Multilingual NMT - Data}
This work does not consider the imbalances present in multilingual NMT.
In real-world scenarios, the quantity of available corpora varies significantly across different languages, which might lead to biased or suboptimal results.
However, considering unbalanced scenarios introduces many additional factors, such as different ways of sampling training data.
Therefore, we constrain our focus to the balanced situation in this work, recognizing the need for future research to address this complexity and explore methods that can better handle the imbalances inherent in multilingual contexts.

\paragraph{Imbalance in Multilingual NMT - Reward}
Although we kept the amount of data the same for different language pairs, we still observed training imbalance problems due to ``uneven'' distribution of rewards.
Specifically, in \tablename~\ref{tab:main-results-high}, neither RAFT nor RAFT+ consistently improves NLLB in En$\rightarrow$De and En$\rightarrow$Zh (both are \textit{from-English} direction) at the high resource setting.
On the other hand, the performance of the other two \textit{to-English} directions (De$\rightarrow$En and Zh$\rightarrow$En) has been significantly improved by notable margins (BLEURT | De$\rightarrow$En: 53.5$\rightarrow$71.3, Zh$\rightarrow$En: 48.4$\rightarrow$62.9).
This means that during training, the model allocates more ``capacity'' to the \textit{to-English} directions than to the \textit{from-English} directions, i.e., they are not balanced.
The intuitive reason is that the model ``finds'' optimizing the \textit{to-English} directions offers a quick boost in reward, thereby ``ignoring'' the \textit{from-English} directions which have less room for improvement.
This involves \textit{how the reward can be reasonably distributed among different language pairs}, which is less discussed in this work, but a meaningful and underexplored future direction.

\paragraph{Applicability of the Method to Mitigate Overoptimization}
The current method for mitigating overoptimization focuses on detecting relatively easy-to-identify errors such as len-ratio and off-target errors.
More elusive mistakes, such as hallucinations, remain unaddressed and might potentially lead to overoptimization as well~\cite{guerreiro-etal-2023-looking}.
In addition, current language detectors are not reliable for extremely low-resource languages~\cite{aji-etal-2022-one}, limiting the applicability in these contexts.

\bibliography{anthology,custom}

\appendix
\clearpage
\section{Error Detection and Penalty}
\label{sec:appendix-error-detection}
We penalize for len-ratio errors and off-target errors.
We consider translations whose length ratios do not belong to $[L,U]$ as len-ratio errors.
We computed the distribution of length ratios of the SFT training data for each language pair (using the tokenizer of the corresponding model), taking $[L,U]$ so that it covers the top 50\% of the most frequent length ratios.
We adopt language detector from~\citet{Stahl2023} to detect off-target errors.

\section{MRT Implementation}
\label{sec:appendix-mrt-implementation}
MRT~\cite{shen-etal-2016-minimum} is designed to directly optimize the model with respect to evaluation metrics.
However, we failed to get meaningful improvements when using MRT.

The \textit{risk} of MRT can be formulate as:
\begin{equation}
\mathcal{R}(\theta)=\sum_{\left(x, y^*\right) \in \mathcal{D}} \mathbb{E}_{y \sim P(y | x ; \theta)}\left[\Delta\left(y, y^*\right)\right],
\end{equation}
where $y^*$ represents the ground-truth translation.
The $\Delta$ function serves as the loss function and can be instantiated using various evaluation metrics, for example, $1 - \mathrm{BLEU}(y, y^*)$.
The objective or MRT is to minimize $\mathcal{R}(\theta)$.

In out setting, $\mathcal{D}$ is the training distribution of $x$, i.e., monolingual data, and we use QE-based reward $r(x, y)$ to indicate the quality of $y$ as the translation of $x$.
Therefore, we modify the \textit{risk} as:
\begin{equation}
\mathcal{R}(\theta)=\sum_{x \in \mathcal{D}} \mathbb{E}_{y \sim P(y | x ; \theta)}\left[ 1 - r(x, y)\right].
\end{equation}

In practice, we sample $k$ candidates and normalize the probabilities to approximate the expectation.
Let $S_k$ denote a set containing $k$ candidates drawn from the distribution $P_T(y | x ; \theta)$, where $T$ is the sampling temperature.
$R(\theta)$ is approximated as:
\begin{equation}
\tilde{R}(\theta)=\sum_{x \in D} \sum_{y \in S_k} \tilde{P}(y | x ; \theta)[1-r(x, y)],
\end{equation}
where
\begin{equation}
\tilde{P}(y | x ; \theta) = \frac{P(y | x ; \theta)^\alpha}{\sum_{y^{\prime} \in S_k}P(y^{\prime} | x ; \theta)^\alpha},
\end{equation}
and $\alpha$ is a hyperparameter. 
Algorithm \ref{alg:mrt} shows the details of MRT.
Similar to RAFT+, when adding a penalty term to reward, we call it MRT+.
We set the penalty term $P=1$ and $\alpha=0.005$\footnote{We follow the default setting in~\citet{sennrich-etal-2017-nematus}.}, and leave other hyperparameters the same as RAFT.
\begin{algorithm}[htpb]
    \caption{MRT}\label{alg:mrt}
    \begin{algorithmic}[1]
        \Require Training set $\mathcal{X}$, reward function $r(x, y)$, initial model $M_0=P(y | x; \theta_0)$, batch size $b$, temperature $T$, the number of candidate $k$
        \For{iteration $i$ in ${0,1,\dots, N-1}$}
            \State $D_i \gets \text{SampleBatch}(\mathcal{X}, b)$
            \State $\mathcal{B}=\emptyset$
            \For{$x \in D_i$}
                \State $y_1,\dots,y_k \sim P_T(y|x;\theta_i)$
                \State $S_k = \{y_1,\dots,y_k\}$
                \State $\mathcal{B}=\mathcal{B}\cup\{(x, S_k)\}$
            \EndFor
            \State Fine-tune $\theta_i$ on $\mathcal{B}$ to obtain $M_{i+1}=P(y | x; \theta_{i+1})$ using MRT risk: $\tilde{R}({\theta}_i)$.
        \EndFor    
    \end{algorithmic}
\end{algorithm}
\begin{figure}[htpb!]
    \centering
    \includegraphics[width=0.98\linewidth]{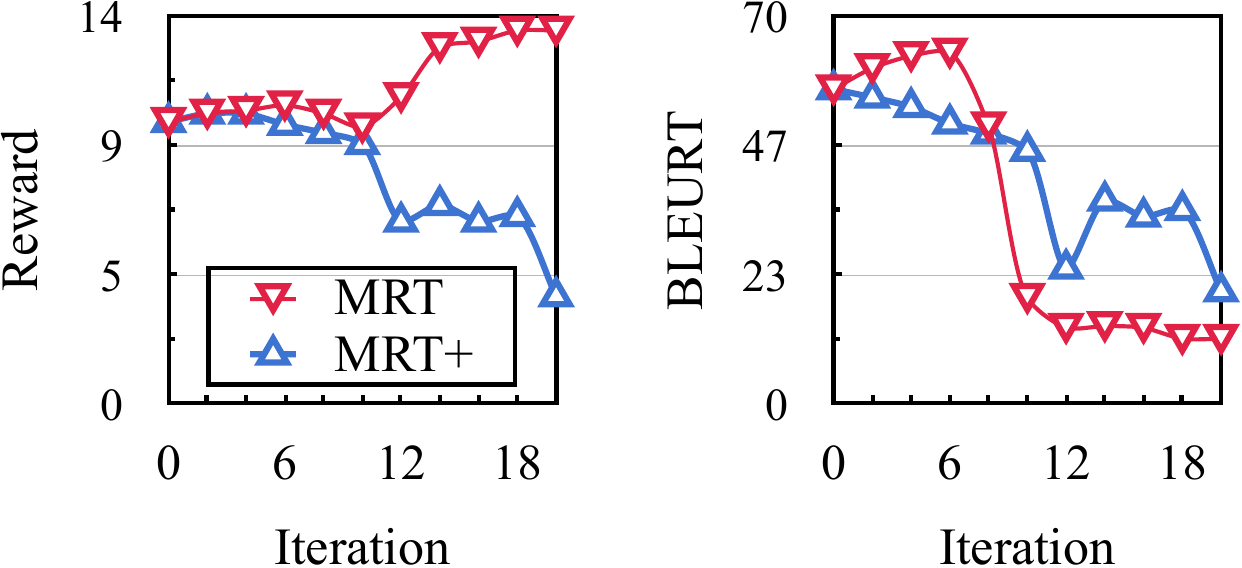}
    \caption{Training curves of MRT and MRT+ under high-resource setting. NLLB-200-1.3B and \textsc{Comet-qe-mqm} are used as base model and QE-based reward model, respectively.}
    \label{fig:mrt-training-curves}
\end{figure}

\figurename~\ref{fig:mrt-training-curves} shows that vanilla MRT suffers from the overoptimization problem, manifested as an increase in reward while translation quality declines.
Additionally, MRT+ also poses challenges in achieving stable convergence.

\section{More Training Curves}
\label{sec:appendix-more-training-curves}
\begin{figure*}[t!]
    \centering
    \begin{subfigure}{\textwidth}
        \centering
        \includegraphics[width=\linewidth]{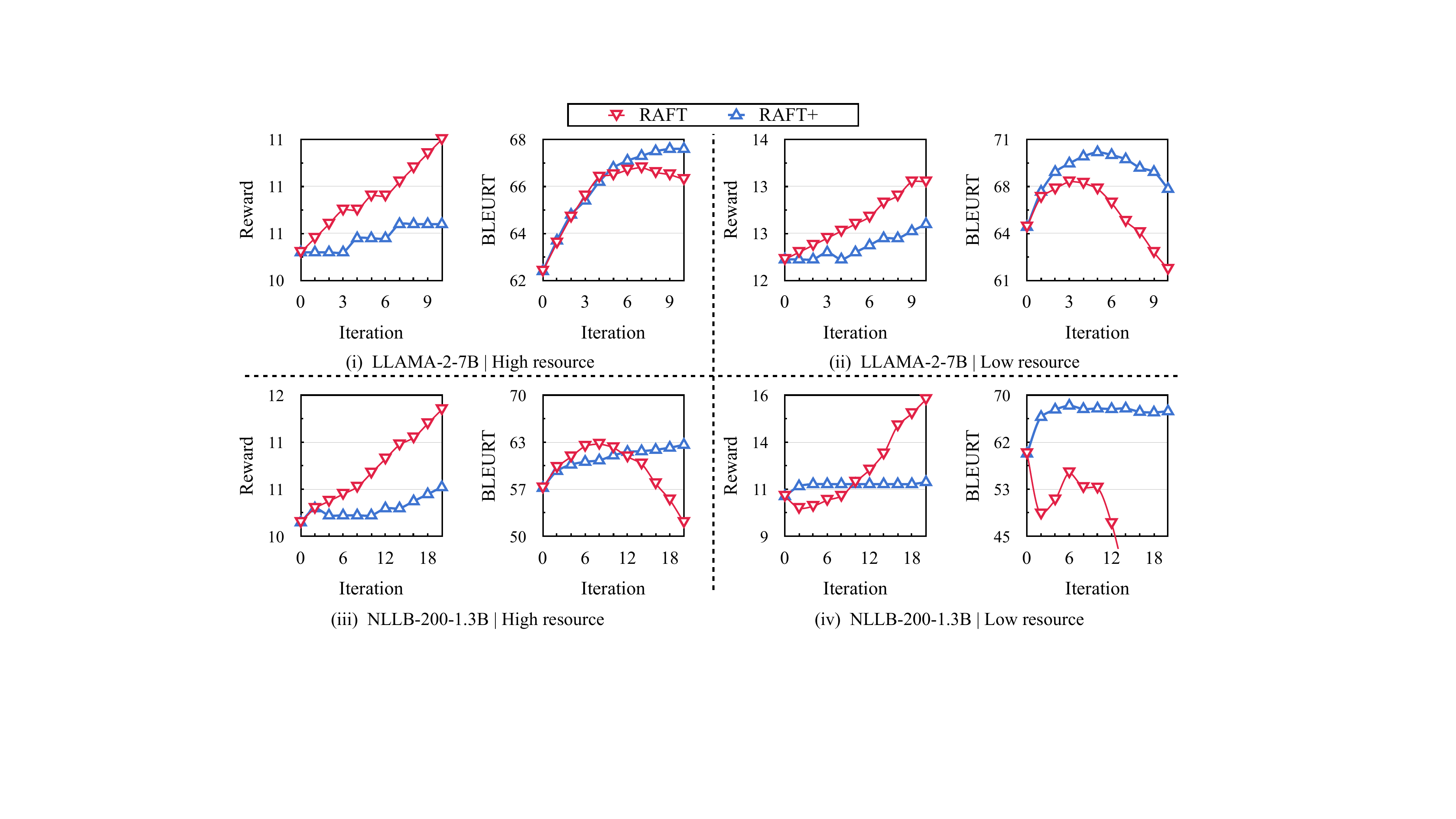}
        \caption{\textsc{Comet-qe-mqm}}
        \label{fig:sub1}
    \end{subfigure}
    
    \vspace{10pt} 

    \begin{subfigure}{\textwidth}
        \centering
        \includegraphics[width=\linewidth]{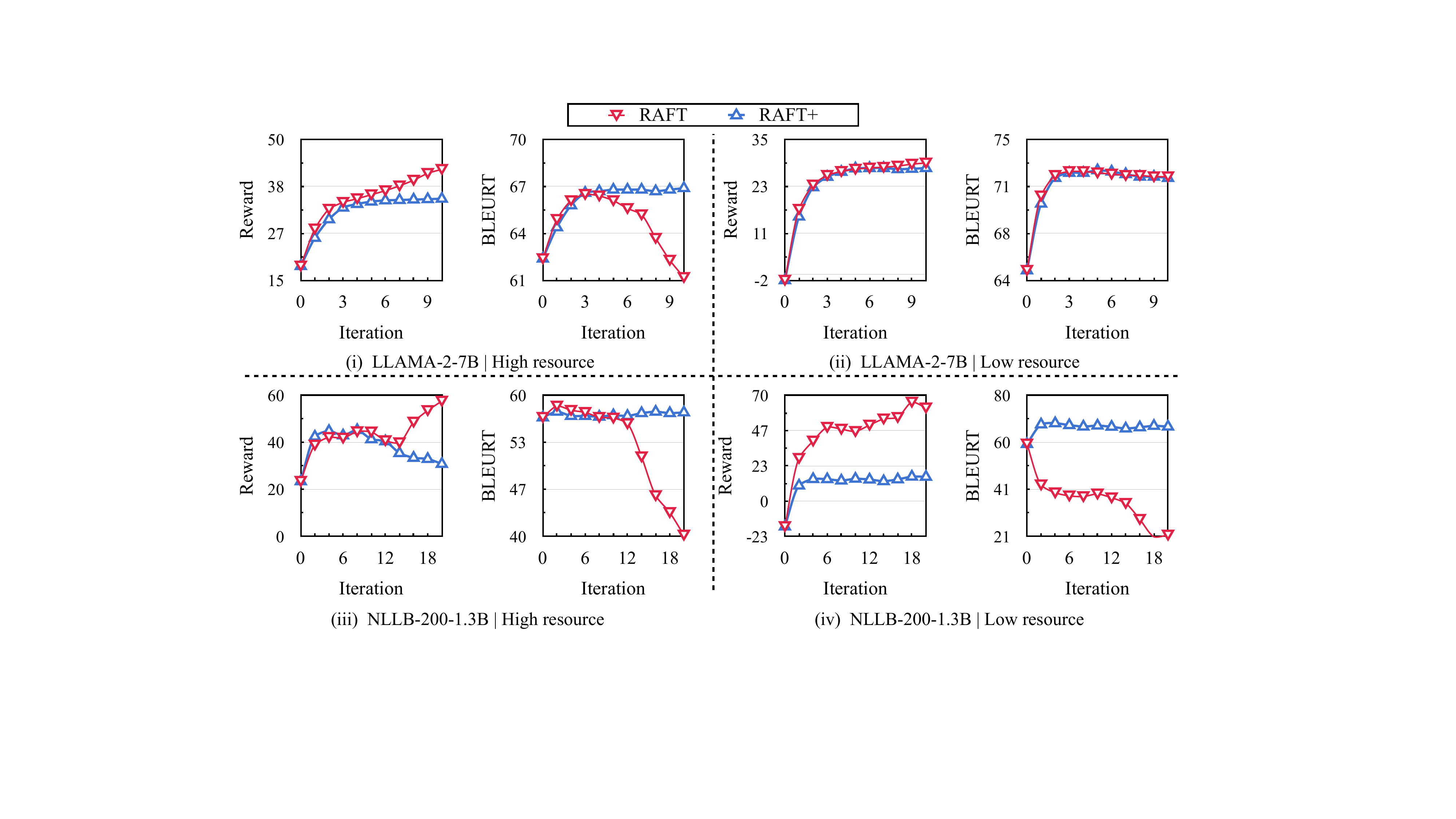}
        \caption{\textsc{UniTE-MUP}}
        \label{fig:sub2}
    \end{subfigure}
    \caption{Training curves under various settings when using \textsc{Comet-qe-mqm} and \textsc{UniTE-MUP} as reward models. The metrics are average values for all language pairs on the development set.}
    \label{fig:more-training-curves}
\end{figure*}

\figurename~\ref{fig:more-training-curves} shows the training curves when using \textsc{Comet-qe-mqm} and \textsc{UniTE-MUP} as the reward models.
Consistent with \figurename~\ref{fig:training-curve}, vanilla RAFT suffers from severe overoptimization problems in most cases, which are greatly alleviated by RAFT+.

\section{\textsc{UniTE-MUP} as the Reward Model}
\label{sec:appendix-unite-mup-as-the-reward-model}
\tablename~\ref{tab:unite-mup-results} presents the test results when \textsc{UniTE-MUP} functions as the reward model.
The main trends are consistent with those shown in Table~\ref{tab:main-results}.
We also observe that neither RAFT nor RAFT+ achieves positive improvement in high-resource language pairs when \textsc{Nllb-200-1.3B} is the base model.
We speculate that this lack of improvement is due to overoptimization problems caused by the presence of errors that are not related to length ratio and off-target errors.
\begin{table*}[htpb]
    \centering
    \begin{subtable}{\textwidth}
    \centering
    \resizebox{0.8\linewidth}{!}{
    \begin{tabular}{l ccccccccll}
        \toprule
        \multirow{2}{*}{ Method} & \multicolumn{2}{c}{ De$\Rightarrow$En} & \multicolumn{2}{c}{ En$\Rightarrow$De} & \multicolumn{2}{c}{ Zh$\Rightarrow$En} & \multicolumn{2}{c}{ En$\Rightarrow$Zh} & \multicolumn{2}{c}{\bf Average} \\
        \cmidrule(lr){2-3}\cmidrule(lr){4-5}\cmidrule(lr){6-7}\cmidrule(lr){8-9}\cmidrule(lr){10-11}
        & {\small COMET} & {\small BLEURT}   & {\small COMET} & {\small BLEURT} & {\small COMET} & {\small BLEURT} & {\small COMET} & {\small BLEURT} & {\small\bf COMET} & {\small\bf BLEURT} \\
        \midrule
        \multicolumn{11}{c}{\textsc{ Llama-2-7B}}    \\
        \textsc{SFT}     &   82.5 &   70.5 &   80.7 &   68.2 &   76.1 &   62.3 &   84.9 &   69.3 &   81.0            &   67.6\\
        \midrule
        \textsc{RAFT}    &   83.3 &   71.6 &   82.8 &   71.4 &   78.3 &   64.8 &   85.4 &   69.6 &   82.4\ii{1.4}    &   69.4\ii{1.8}\\
        \textsc{RAFT+}   &   83.4 &   71.6 &   84.2 &   73.6 &   78.9 &   66.1 &   85.0 &   69.0 &\bf82.9\ii{\bf1.9} &\bf70.1\ii{\bf2.5}\\
        \midrule
        \midrule
        \multicolumn{11}{c}{\textsc{ Nllb-200-1.3B}}    \\
        \textsc{SFT}     &   70.9 &   52.5 &   85.3 &   74.8 &   66.0 &   48.4 &   83.7 &   69.1 &   76.5            &\bf61.2\\
        \midrule
        \textsc{RAFT}    &   72.7 &   50.9 &   85.8 &   75.4 &   66.5 &   48.6 &   84.5 &   69.1 &\bf77.4\ii{\bf0.9} &   61.0\dd{0.2}\\
        \textsc{RAFT+}   &   72.8 &   50.3 &   85.8 &   75.5 &   65.5 &   47.1 &   84.4 &   69.0 &   77.1\ii{0.6}    &   60.5\dd{0.7}\\
        \bottomrule
    \end{tabular}}
    \caption{High-resource language pairs}
    \label{tab:unite-mup-results-high}
    \end{subtable}

    \vspace{10pt}

    \begin{subtable}{\textwidth}
    \centering
    \resizebox{0.8\linewidth}{!}{
    \begin{tabular}{l ccccccccll}
        \toprule
        \multirow{2}{*}{ Method} & \multicolumn{2}{c}{ En$\Rightarrow$Uk} & \multicolumn{2}{c}{ Uk$\Rightarrow$En} & \multicolumn{2}{c}{ Uk$\Rightarrow$Cs} & \multicolumn{2}{c}{ Cs$\Rightarrow$Uk} & \multicolumn{2}{c}{\bf Average} \\
        \cmidrule(lr){2-3}\cmidrule(lr){4-5}\cmidrule(lr){6-7}\cmidrule(lr){8-9}\cmidrule(lr){10-11}
        & {\small COMET} & {\small BLEURT}   & {\small COMET} & {\small BLEURT} & {\small COMET} & {\small BLEURT} & {\small COMET} & {\small BLEURT} & {\small\bf COMET} & {\small\bf BLEURT} \\
        \midrule
        \multicolumn{11}{c}{\textsc{ Llama-2-7B}}    \\
        \textsc{SFT}     &   79.2 &   64.0 &   76.7 &   66.0 &   70.0 &   53.2 &   71.2 &   51.3 &   74.3            &   58.6\\
        \midrule
        \textsc{RAFT}    &   80.9 &   66.4 &   80.7 &   70.2 &   81.6 &   68.9 &   83.6 &   69.1 &   81.7\ii{7.4}    &   68.6\ii{10.0}\\
        \textsc{RAFT+}   &   81.3 &   67.1 &   81.0 &   70.5 &   81.5 &   68.7 &   84.0 &   69.6 &\bf81.9\ii{\bf7.6} &\bf69.0\ii{\bf10.4}\\
        \midrule
        \midrule
        \multicolumn{11}{c}{\textsc{ Nllb-200-1.3B}}    \\
        \textsc{SFT}     &   83.1 &   70.2 &   71.1 &   62.7 &   73.2 &   61.5 &   57.3 &   43.4 &   71.2            &   59.4\\
        \midrule
        \textsc{RAFT}    &   85.1 &   72.4 &   64.5 &   30.9 &   70.5 &   26.9 &   74.1 &   27.4 &   73.5\ii{2.3}    &   39.4\dd{19.2}\\
        \textsc{RAFT+}   &   84.3 &   71.4 &   77.0 &   66.5 &   82.6 &   70.2 &   71.6 &   54.9 &\bf78.9\ii{\bf7.7} &\bf65.8\ii{\bf6.4}\\
        \bottomrule
    \end{tabular}}
    \caption{Low-resource language pairs}
    \label{tab:unite-mup-results-low}
    \end{subtable}
    \caption{Test results under various settings when \textsc{UniTE-MUP} functions as the reward model. Bold indicates that the average performance of the method exceeds that of SFT and RAFT/RAFT+ within the same QE model. The subscripts indicate the change in performance relative to the SFT.}
    \label{tab:unite-mup-results}
\end{table*}

\section{chrF as the Evaluation Metric}
\label{sec:appendix-chrf-as-the-evaluation-metric}
\tablename~\ref{tab:main-results-chrf} shows the chrF values for main results.
\begin{table*}[htpb!]
    \centering
    \begin{subtable}{0.49\linewidth}
    \centering
    \resizebox{1.0\linewidth}{!}{
    \begin{tabular}{l c c c c l}
        \toprule
        Method & De$\Rightarrow$En & En$\Rightarrow$De & Zh$\Rightarrow$En & En$\Rightarrow$Zh & \bf Average \\
        \midrule
        \multicolumn{6}{c}{\textsc{ Llama-2-7B}}    \\
        \textsc{SFT}     &   52.1 &   51.5 &   46.3 &   34.4 &   46.0\\
        \midrule
        \multicolumn{6}{l}{\textsc{Reward model: Comet-qe-da}}    \\[3pt]
        \textsc{~~RAFT}  &   55.4 &   56.1 &   50.2 &   35.5 &   49.3\ii{3.3}\\
        \textsc{~~RAFT+} &   56.1 &   58.8 &   51.7 &   34.5 &   50.3\ii{\bf 4.3}\\
        \midrule
        \multicolumn{6}{l}{\textsc{Reward model: Comet-qe-mqm}}    \\[3pt]
        \textsc{~~RAFT}  &   53.4 &   56.8 &   47.2 &   34.9 &   48.1\ii{2.1}\\
        \textsc{~~RAFT+} &   54.5 &   58.5 &   49.2 &   34.8 &   49.3\ii{\bf 3.3}\\
        \midrule
        \midrule
        \multicolumn{6}{c}{\textsc{ Nllb-200-1.3B}}    \\
        \textsc{SFT}     &   35.3 &   60.4 &   15.2 &   30.7 &   35.4\\
        \midrule
        \multicolumn{6}{l}{\textsc{Reward model: Comet-qe-da}}    \\[3pt]
        \textsc{~~RAFT}  &   35.1 &   60.8 &   22.8 &   30.9 &   37.4\ii{2.0}\\
        \textsc{~~RAFT+} &   42.6 &   60.6 &   27.8 &   30.7 &   40.4\ii{\bf 5.0}\\
        \midrule
        \multicolumn{6}{l}{\textsc{Reward model: Comet-qe-mqm}}    \\[3pt]
        \textsc{~~RAFT}  &   48.3 &   60.7 &   35.5 &   30.7 &   43.8\ii{8.4}\\
        \textsc{~~RAFT+} &   49.7 &   60.6 &   43.2 &   30.6 &   46.1\ii{\bf 10.7}\\
        \bottomrule
    \end{tabular}}
    \caption{High-resource language pairs}
    \label{tab:main-results-high-chrf}
    \end{subtable}
    \hfill
    \begin{subtable}{0.49\linewidth}
    \centering
    \resizebox{1.0\linewidth}{!}{
    \begin{tabular}{l c c c c l}
        \toprule
        Method & En$\Rightarrow$Uk & Uk$\Rightarrow$En & Uk$\Rightarrow$Cs & Cs$\Rightarrow$Uk & \bf Average \\
        \midrule
        \multicolumn{6}{c}{\textsc{ Llama-2-7B}}    \\
        \textsc{SFT}     &   43.1 &   51.1 &   28.3 &   30.2 &   38.2\\
        \midrule
        \multicolumn{6}{l}{\textsc{Reward model: Comet-qe-da}}    \\[3pt]
        \textsc{~~RAFT}  &   46.2 &   56.4 &   43.9 &   47.5 &   48.5\ii{10.3}\\
        \textsc{~~RAFT+} &   46.5 &   56.9 &   44.5 &   48.0 &   49.0\ii{\bf 10.8}\\
        \midrule
        \multicolumn{6}{l}{\textsc{Reward model: Comet-qe-mqm}}    \\[3pt]
        \textsc{~~RAFT}  &   40.6 &   48.1 &   30.0 &   30.3 &   37.3\dd{0.9}\\
        \textsc{~~RAFT+} &   44.3 &   53.5 &   36.1 &   38.2 &   43.0\ii{\bf 4.8}\\
        \midrule
        \midrule
        \multicolumn{6}{c}{\textsc{ Nllb-200-1.3B}}    \\
        \textsc{SFT}     &   50.6 &   45.3 &   36.6 &   25.7 &   39.5\\
        \midrule
        \multicolumn{6}{l}{\textsc{Reward model: Comet-qe-da}}    \\[3pt]
        \textsc{~~RAFT}  &   52.1 &   10.2 &   10.2 &   10.3 &   20.7\dd{18.8}\\
        \textsc~~{RAFT+} &   51.2 &   55.1 &   49.7 &   43.3 &   49.8\ii{\bf 10.3}\\
        \midrule
        \multicolumn{6}{l}{\textsc{Reward model: Comet-qe-mqm}}    \\[3pt]
        \textsc{~~RAFT}  &   51.7 &   31.5 &   22.5 &   26.0 &   33.0\dd{6.5}\\
        \textsc{~~RAFT+} &   51.5 &   54.1 &   48.0 &   41.5 &   48.8\ii{\bf 9.3}\\
        \bottomrule
    \end{tabular}}
    \caption{Low-resource language pairs}
    \label{tab:main-results-low-chrf}
    \end{subtable}
    \caption{chrF results of \tablename~\ref{tab:main-results}}
    \label{tab:main-results-chrf}
\end{table*}
\end{document}